\documentclass[10pt,twocolumn,letterpaper]{article}
\usepackage[table,dvipsnames]{xcolor}
\usepackage{tikz}  
\usepackage{cvpr}
\usepackage{times}
\usepackage{graphicx}
\usepackage{amsmath}
\usepackage{amssymb}
\usepackage{textcomp}

\usepackage{subfiles}
\usepackage{pgfplots} 
\usepackage{pgfplotstable} 
\pgfplotsset{compat=newest} 
\usetikzlibrary{plotmarks} 
\usetikzlibrary{colorbrewer} 
\usepackage{booktabs} 
\usepackage{etoolbox,siunitx} 

\robustify\bfseries
\robustify\itshape
\usepackage{paralist} 

\usepackage{flushend}

\definecolor{olivegreen}{RGB}{0,170,0}
\definecolor{darkred}{RGB}{220,100,10}
\definecolor{tealblue}{RGB}{20,100,200}

\newcommand{\ofl}{Tsai2016}
\newcommand{\bvs}{NicolasMaerki2016}
\newcommand{\fcp}{Perazzi2015}
\newcommand{\jmp}{Fan2015}
\newcommand{\hvs}{Grundmann2010}
\newcommand{\sea}{Ramakanth2014}
\newcommand{\tsp}{Chang2013}
\newcommand{\ltv}{Ochs2014}
\newcommand{\hbt}{Godec2013}
\newcommand{\afs}{Vijayanarasimhan2012}
\newcommand{\jfs}{ShankarNagaraja2015}

\newcommand{\mtk}{Kho+17}

\newcommand{\fst}{Papazoglou2013}
\newcommand{\sal}{Shen2015}
\newcommand{\key}{Lee2011}
\newcommand{\msg}{Brox2010}
\newcommand{\trc}{Fragkiadaki2012}
\newcommand{\cvos}{Taylor2015}
\newcommand{\nlc}{Faktor2014}
\newcommand{\scf}{Jain2014}

\newcommand{\mcg}{Pont-Tuset2016}
\newcommand{\cob}{Maninis2016}

\newcommand{\ours}{OSVOS}
\newcommand{\longours}{One-Shot Video Object Segmentation}

\newcommand{\nofl}{OFL}
\newcommand{\nbvs}{BVS}
\newcommand{\nfcp}{FCP}
\newcommand{\njmp}{JMP}
\newcommand{\nhvs}{HVS}
\newcommand{\nsea}{SEA}
\newcommand{\ntsp}{TSP}
\newcommand{\nltv}{LTV}
\newcommand{\nhbt}{HBT}
\newcommand{\nafs}{AFS}
\newcommand{\nscf}{SCF}
\newcommand{\njfs}{JFS}

\newcommand{\nfst}{FST}
\newcommand{\nsal}{SAL}
\newcommand{\nkey}{KEY}
\newcommand{\nmsg}{MSG}
\newcommand{\ntrc}{TRC}
\newcommand{\ncvos}{CVOS}
\newcommand{\nnlc}{NLC}

\newcommand{\nmcg}{MCG}
\newcommand{\ncob}{COB}

\newcommand{\J}{\mathcal{J}}
\newcommand{\F}{\mathcal{F}}
\newcommand{\T}{\mathcal{T}}

\definecolor{rowblue}{RGB}{220,230,240}

\usepackage[pagebackref=true,breaklinks=true,letterpaper=true,colorlinks,bookmarks=false]{hyperref}

\cvprfinalcopy 

\ifcvprfinal\fi
\begin{document}

\title{One-Shot Video Object Segmentation}

\author{S. Caelles$^{1,}$\thanks{First two authors contributed equally} \quad
	K.-K. Maninis$^{1,*}$\quad
	J. Pont-Tuset$^1$ \quad
	L. Leal-Taix\'e$^2$\quad
    D. Cremers$^2$\quad
	L. Van Gool$^{1}$\\
$^1$ETH Z\"urich \quad $^2$TU M\"unchen
}

\maketitle

\vspace{2mm}
\begin{abstract}
\vspace{-1mm}
This paper tackles the task of semi-supervised video object segmentation, \ie,
the separation of an object from the background in a video, given the mask of the first frame.
We present \longours{} (\ours{}), based on a fully-convolutional neural network architecture that
is able to successively transfer generic semantic information, learned on ImageNet, to the task of foreground segmentation, and finally to learning the appearance of a single annotated object of the test sequence (hence one-shot).
Although all frames are processed independently, the results are temporally coherent and stable.
We perform experiments on two annotated video segmentation databases, which show that \ours{} is
fast and improves the state of the art by a significant margin (79.8\% vs 68.0\%).
\end{abstract}


\section{Introduction}
\label{sec:intro}

\subsection*{From Pre-Trained Networks...}
Convolutional Neural Networks (CNNs) are revolutionizing many fields of computer vision. For instance, they have dramatically boosted the performance for problems like image classification~\cite{Krizhevsky2012,SiZi15,He+16} and object detection~\cite{Girshick2014,Gir15,Liu+16}. Image segmentation has also been taken over by CNNs recently~\cite{Maninis2016,Kokkinos2016,XiTu15,BST15b,BST16},
with deep architectures
pre-trained on the weakly related task of image classification on ImageNet~\cite{Russakovsky2015}. One of the major downsides of deep network approaches is their hunger for training data. Yet, with various pre-trained network architectures one may ask how much training data do we really need for the specific problem at hand? This paper investigates segmenting an object along an entire video, when we only have one single labeled training example, e.g. the first frame.

\subsection*{...to \longours{}}
This paper presents \textit{\longours{}} (\ours{}), a CNN architecture to
tackle the problem of semi-supervised video object segmentation, that is,
the classification of all pixels of a video sequence into background and foreground, given
the manual annotation of one (or more) of its frames.
Figure~1 shows an example result of \ours{}, where the input is the segmentation of the first frame
(in red), and the output is the mask of the object in the 90 frames of the sequence (in green).

\begin{figure*}
\centering
\vspace{-2mm}
\includegraphics[width=0.8\textwidth]{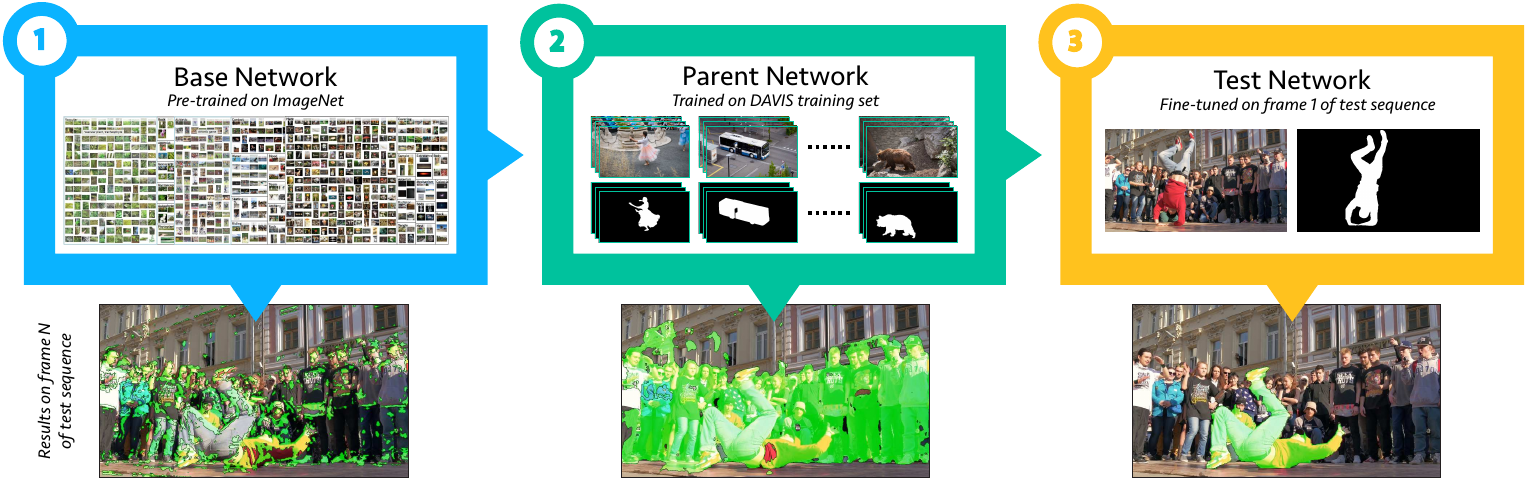}\\[1mm]
\caption{\textbf{Overview of \ours{}}: (1) We start with a pre-trained base CNN for image labeling on ImageNet; its results in terms of segmentation, although conform with some image features, are not useful.
(2) We then train a \textit{parent network} on the training set of DAVIS; the segmentation results improve but are not focused on an specific object yet. (3) By fine-tuning on a segmentation example for the specific target object in a single frame, the network rapidly focuses on that target.}
\label{fig:overview}
\vspace{-2mm}
\end{figure*}

The first contribution of the paper is to adapt the CNN to a particular object instance
given a single annotated image (hence \textit{one-shot}).
To do so, we adapt a CNN pre-trained on image recognition~\cite{Russakovsky2015} to video object segmentation. This is achieved by training it on a set of videos with manually segmented objects. Finally, it is  fine-tuned \textit{at test time} on a specific object that is manually segmented in a single frame.
Figure~\ref{fig:overview} shows the overview of the method.
Our proposal tallies with the observation that 
leveraging these different levels of information
to perform object segmentation would stand to reason: from generic semantic information of a large
amount of categories, passing through the knowledge of the \textit{usual} shapes of objects,
down to the specific properties of a particular object we are interested in segmenting.

The second contribution of this paper is that \ours{} 
processes each frame of a video independently, obtaining temporal consistency as a by-product rather
than as the result of an explicitly imposed, expensive constraint.
In other words, we cast video object segmentation as a per-frame segmentation problem given
the \textit{model} of the object from one (or various) manually-segmented frames.
This stands in contrast to the dominant approach where temporal consistency plays the central role, assuming that objects do not change too much between one frame and the next. Such methods adapt their single-frame models smoothly throughout the video, looking for targets whose shape and appearance vary \textit{gradually} in consecutive frames, but fail when those constraints do not apply, unable to recover from relatively common situations such as occlusions and abrupt motion.

In this context, {\it motion} estimation has emerged as a key ingredient for state-of-the-art video segmentation 
algorithms~\cite{Tsai2016,Ramakanth2014,Grundmann2010}. Exploiting it is not a trivial task however, as one \eg. has to compute temporal matches in the form of optical flow or dense trajectories~\cite{Brox2010},
which can be an even harder problem. 

We argue that temporal consistency was needed in the past, as one had to overcome major drawbacks of the then inaccurate shape or appearance models.
On the other hand, in this paper deep learning will be shown to provide a sufficiently accurate model of the target object to produce temporally stable results even when processing each frame independently.
This has some natural advantages: \ours{} is able to segment objects through occlusions, it is not limited to certain ranges of motion, it does not need to process frames sequentially, and errors are not temporally propagated. In practice, this allows \ours{} to handle \eg interlaced videos of surveillance scenarios, where cameras can go blind for a while before coming back on again.

Our third contribution is that \ours{} can work at various points of the trade-off between
speed and accuracy.
In this sense, it can be adapted in two ways.
First, given one annotated frame, the user can choose the level of fine-tuning of \ours{}, giving him/her the freedom between a faster method or more accurate results.
Experimentally, we show that \ours{} can run at 181 ms per frame and 71.5\% accuracy, and up to 79.7\% when processing each frame in 7.83 s.
Second, the user can annotate more frames, those on which the current segmentation is less satisfying, upon which \ours{} will refine the result. 
We show in the experiments that the results indeed improve gradually with more supervision, reaching
an outstanding level of 84.6\% with two annotated frames per sequence, and 86.9\% with four, from 79.8\% from one annotation.

Technically, we adopt the architecture of Fully Convolutional Networks (FCN)~\cite{farabet2013learning, LSD15}, suitable for dense predictions. FCNs have recently become popular due to their performance both in terms of accuracy and computational efficiency~\cite{LSD15,Dai2016a,Dai2016}.
Arguably, the Achilles' heel of FCNs when it comes to segmentation is the coarse scale of the deeper layers, which leads to inaccurately localized predictions.
To overcome this, a large variety of works from different fields use skip connections of larger feature maps~\cite{LSD15,Har+15,XiTu15,Man+16}, 
or learnable filters to improve upscaling~\cite{NHH15,Yan+16}.
To the best of our knowledge, this work is the first to use FCNs for the task of video segmentation. 

We perform experiments on two video object segmentation datasets (DAVIS~\cite{Perazzi2016} and Youtube-Objects~\cite{Prest2012,Jain2014}) and show that \ours{} significantly improves the state of the art 79.8\% vs 68.0\%.
Our technique is able to process a frame of DAVIS (480$\times$854 pixels) in 102 ms. By increasing the level of supervision, \ours{} can further improve its results to 86.9\% with just four annotated frames per sequence, thus providing a vastly accelerated rotoscoping tool.

All resources of this paper, including training and testing code, pre-computed results, and pre-trained models are publicly available at \url{www.vision.ee.ethz.ch/~cvlsegmentation/osvos/}.

\section{Related Work}
\label{sec:related}

\paragraph*{Video Object Segmentation and Tracking:}
Most of the current literature on semi-supervised video object segmentation enforces 
temporal consistency in video sequences to propagate the initial mask into the following frames.
First of all, in order to reduce the computational complexity some works make use of superpixels~\cite{\tsp,\hvs}, patches~\cite{\sea,\jmp}, or even object proposals~\cite{\fcp}.
M\"arki \etal~\cite{\bvs} cast the problem into a bilateral space in order to solve it more efficiently.
After that, an optimization using one of the previous aggregations of pixels is usually performed; which can consider the full video sequence~\cite{\fcp,\bvs}, a subset of frames~\cite{\hvs}, or only the results in frame $n$ to obtain the mask in $n+1$~\cite{\sea,\tsp,\jmp}.
As part of their pipeline, some of the methods include the computation of optical flow~\cite{\hvs,\sea}, which considerably reduces speed.
Concurrent works have also used deep learning to address Video Object Segmentation. MaskTrack~\cite{\mtk} learns to refine the detected masks frame by frame, by using the detections of the previous frame, along with Optical Flow and post-processing with CRFs. In~\cite{JGG17}, the authors combine training of a CNN with ideas of bilateral filtering. Different from those approaches, OSVOS is a simpler pipeline which segments each frame independently, and produces more accurate results, while also being significantly faster.

In the case of visual tracking (bounding boxes instead of segmentation) Nam and Han~\cite{Nam2016} use a CNN to learn a representation of the object to be tracked, but only to look for the most similar window in frame $n+1$ given the object in frame $n$. In contrast, our CNN learns a single model from frame 1 and segments the rest of the frames from this model. 

\vspace{-3mm}
\paragraph*{FCNs for Segmentation:} 
Segmentation research has closely followed the innovative ideas of CNNs in the last few years.
The advances observed in image recognition~\cite{Krizhevsky2012, SiZi15, He+16}
have been beneficial to segmentation in many forms (semantic~\cite{LSD15, NHH15}, instance-
level~\cite{Gir15,Pin+16,Dai2016a}, biomedical~\cite{RFB15}, generic~\cite{Maninis2016}, etc.).
Many of the current best performing methods have in common a deep architecture, usually pre-trained on 
ImageNet, trainable end-to-end.
The idea of dense predictions with CNNs was pioneered by~\cite{farabet2013learning} and formulated 
by~\cite{LSD15} in the form of Fully Convolutional Networks (FCNs) for semantic segmentation.
The authors noticed that by changing the last fully connected layers to $1\times1$ convolutions
it is possible to train on images of arbitrary size,
by predicting correspondingly-sized outputs.
Their approach boosts efficiency over patch-based approaches where one needs to perform redundant
computations in overlapping patches.
More importantly, by removing the parameter-intensive fully connected layers,
the number of trainable parameters drops significantly, facilitating training with
relatively few labeled data.

In most CNN architectures~\cite{Krizhevsky2012, SiZi15, He+16},
activations of the intermediate layers gradually decrease in size,
because of spatial pooling operations or convolutions with a stride.
Making dense predictions from downsampled activations results in coarsely localized outputs~\cite{LSD15}. 
Deconvolutional layers that learn how to upsample are used in~\cite{NHH15,Yan+16}.
In~\cite{Pin+16}, activations from shallow layers are gradually
injected into the prediction to favor localization.
However, these architectures come with many more trainable parameters and their use is limited to cases with sufficient data.

Following the ideas of FCNs, Xie and Tu~\cite{XiTu15} separately supervised the intermediate layers of a deep 
network for contour detection.
The duality between multiscale contours and hierarchical segmentation~\cite{Arb+11,Pont-Tuset2016}
was further studied by Maninis \etal~\cite{Maninis2016} by bringing CNNs to the field of
generic image segmentation.
In this work we explore how to train an FCN for accurately localized dense prediction
based on very limited annotation: a single segmented frame.

\section{One-Shot Deep Learning}
\label{sec:method}

Let us assume that one would like to segment an object in a video, for which the only available piece of 
information is its foreground/background segmentation in one frame.
Intuitively, one could analyze the entity, create a \textit{model},
and search for it in the rest of the frames.
For humans, this very limited amount of information is more than enough, and changes in appearance, shape, 
occlusions, etc.\ do not pose a significant challenge, because we leverage strong priors: first ``It is an object,'' and then ``It is \textit{this particular} object.''
Our method is inspired by this gradual refinement.

We train a Fully Convolutional Neural Network (FCN) for the binary classification task of separating the foreground object from the background.
We use two successive training steps: First we train on a large variety of objects, offline, to construct a 
model that is able to discriminate the general notion of a foreground object, \ie, ``It is an object.''
Then, at test time, we fine-tune the network for a small number of iterations on the particular
instance that we aim to segment, \ie, ``It is \textit{this particular} object.''
The overview of our method is illustrated in Figure~\ref{fig:overview}.

\subsection{End-to-end trainable foreground FCN}
Ideally, we would like our CNN architecture to satisfy the following criteria:
\begin{compactenum}
\item Accurately localized segmentation output, as discussed in Section~\ref{sec:related}.
\item Relatively small number of parameters to train from a limited amount of annotation data.
\item Relatively fast testing times.
\end{compactenum}
We draw inspiration from the CNN architecture of~\cite{Man+16}, originally used for biomedical
image segmentation.
It is based on the VGG~\cite{SiZi15} network, modified for accurately localized dense prediction (Point 1).
The fully-connected layers needed for classification are removed (Point 2), and efficient image-to-image 
inference is performed (Point 3).
The VGG architecture consists of groups of convolutional plus Rectified Linear Units (ReLU)
layers grouped into 5 stages.
Between the stages, pooling operations downscale the feature maps as we go deeper into the network.
We connect convolutional layers to form separate skip paths from the last layer of each stage
(before pooling).
Upscaling operations take place wherever necessary, and feature maps from the separate paths are
concatenated to construct a volume with information from different levels of detail.
We linearly fuse the feature maps to a single output which has the same dimensions as the image,
and we assign a loss function to it.
The proposed architecture is shown in Figure~\ref{fig:two-stream} (1), foreground branch.

The pixel-wise cross-entropy loss for binary classification (we keep the notation of Xie and Tu~\cite{XiTu15}) is in this case defined as:

{\footnotesize
\begin{equation}
\begin{split}
\mathcal{L}\left(\mathbf{W}\right)\!&=\!-\!\!\sum_{j}\!{y_j\!\log{\!P\left(y_j\!\!=\!\!1 |X;\!\mathbf{W}\right)}\!+\!\left( 1\!\!-\!\!y_j\right)\!\log{\left(1\!\!-\!\!P\left(y_j\!\!=\!\!1 |X;\!\mathbf{W}\right)\!\right)}}\\
&=\!-\!\!\!\sum_{j \in Y_+}\!\!{\!\log{\!P\left(y_j\!\!=\!\!1 |X;\!\mathbf{W}\right)}}-\!\!\sum_{j	\in Y_-}\!\!{\!\log{\!P\left(y_j\!\!=\!\!0 |X;\mathbf{W}\right)}} \label{eq:cross_entropy}
\end{split}\nonumber
\end{equation}}\\[-4mm]
where $\mathbf{W}$ are the standard trainable parameters of a CNN, $X$ is the input image, $y_j \in {0,1}, j=1,..,|X|$ is the pixel-wise binary label of $X$, and $Y_+$ and $Y_-$ are the positive and negative labeled pixels. $P(\cdot)$ is obtained by applying a sigmoid to the activation of the final layer.

In order to handle the imbalance between the two binary classes, Xie and Tu~\cite{XiTu15} proposed a modified version of the cost function, originally used for contour detection (we drop $\mathbf{W}$ for the sake of readability):

{\footnotesize
\begin{align}
\mathcal{L}_{mod}\!=\!-\beta\!\!\sum_{j \in Y_+}\!\!{\!\log{\!P\left(y_j\!\!=\!\!1 |X\right)}}-(1\!-\!\beta)\!\!\sum_{j\in Y_-}\!\!{\!\log{\!P\left(y_j\!\!=\!\!0 |X\right)}} \label{eq:hed_cost}
\end{align}}%
where $\beta=|Y_-|/|Y|$. Equation~\ref{eq:hed_cost} allows training for imbalanced binary tasks~\cite{Kokkinos2016,XiTu15,Maninis2016,Man+16}.

\subsection{Training details}

\paragraph{Offline training:} 
The base CNN of our architecture~\cite{SiZi15} is pre-trained on ImageNet for image labeling,
which has proven to be a very good initialization to other tasks~\cite{LSD15,XiTu15,Kokkinos2016,Maninis2016,Har+15,Yan+16}.
Without further training, the network is not capable of performing segmentation, 
as illustrated in Figure~\ref{fig:overview} (1). We refer to this network as the \textit{``base network}.''

We therefore further train the network on the binary masks of the training set of DAVIS,
to learn a generic notion of how to segment objects from their background, their usual shapes, etc.
We use Stochastic Gradient Descent (SGD) with momentum 0.9 for 50000
iterations.
We augment the data by mirroring and zooming in.
The learning rate is set to $10^{-8}$, and is gradually decreased.
After offline training, the network learns to segment foreground objects from the background,
as illustrated in Figure~\ref{fig:overview} (2).
We refer to this network as the \textit{``parent network}.''

\paragraph{Online training/testing:} With the parent network available, we can proceed to our main task (\textit{``test network}'' in Figure~\ref{fig:overview} (3)):
Segmenting a particular entity in a video, given the image and the segmentation of the first frame.
We proceed by further training (fine-tuning) the parent network for the particular image/ground-truth pair, and then testing on the entire sequence, using the new weights.
The timing of our method is therefore affected by two times: the fine-tuning time (once per annotated mask)
and the segmentation of all frames (once per frame).
In the former we have a trade-off between quality and time: the more iterations we allow the technique to learn, the better results but the longer the user will have to wait for results.
The latter does not depend on the training time: 
\ours{} is able to segment each 480p frame ($480\times854$) in 102 ms.

\begin{figure}
\setlength{\fboxsep}{0pt}
\resizebox{\linewidth}{!}{%
      \fbox{\includegraphics[width=0.3\textwidth]{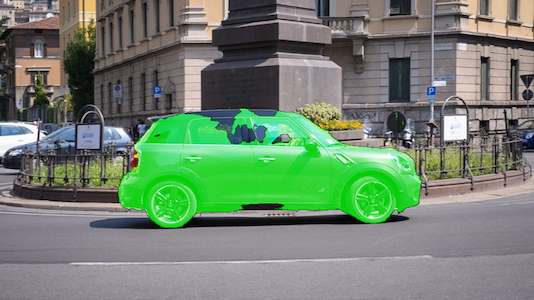}}
      \fbox{\includegraphics[width=0.3\textwidth]{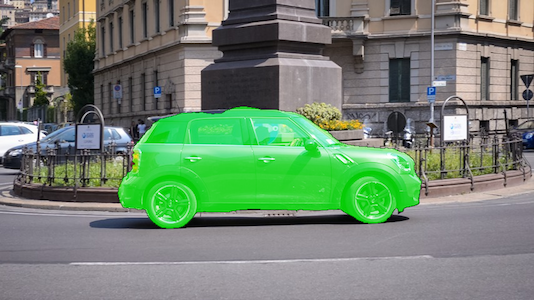}}
      }
\caption{\textbf{Qualitative evolution of the fine tuning}: Results at 10 seconds and 1 minute per sequence.}
\label{fig:learning}
\vspace{-4mm}
\end{figure}

Regarding the fine-tuning time, we present two different modes: One can either need to fine-tune online, by segmenting a frame and waiting for the results in the entire sequence, or offline, having access to the
object to segment beforehand.
Especially in the former mode, there is the need to control the amount of time dedicated to training: the more time allocated for fine-tuning, the more the user waits and the better the results are.
In order to explore this trade-off, in our experiments we train for a period between
10 seconds and 10 minutes per sequence.
Figure~\ref{fig:learning} shows a qualitative example of the evolution of the results' quality depending on the time allowed for fine-tuning.

In the experiments section, Figure~\ref{fig:qual_vs_time} quantifies this evolution.
Ablation analysis shows that both offline and online training are crucial for good performance:
If we perform our online training directly from the ImageNet model, the performance drops significantly. 
Only dropping the online training for a specific object also yields a significantly worse performance, as 
already transpired from Figure~\ref{fig:overview} (2).

\subsection{Contour snapping}
In the field of image classification~\cite{Krizhevsky2012,SiZi15,He+16},
where our base network was designed and trained, spatial invariance is a design choice: no matter where an object appears in the image, the classification result should be the same. 
This is in contrast to the accurate localization of the object contours that we expect in (video) object segmentation.
Despite the use of skip connections~\cite{LSD15,Har+15,XiTu15,Man+16} to minimize the loss of spatial accuracy, we observe that \ours{}'s segmentations have some room for improvement in terms of contour localization. We propose two different strategies to improve the results in this regard.

First, we propose the use of the Fast Bilateral Solver (FBS)~\cite{Barron2016} to snap the background prediction to the image edges.
It performs a Gaussian smoothing in the five-dimensional \textit{color-location} space,
which results in a smoothing of the input signal (foreground segmentation) that preserves the edges of the image.
It is useful in practice because it is fast ($\approx$60 ms per frame),
and it is differentiable so it can be included in an end-to-end trainable deep learning architecture.
The drawback of this approach, though, is that it preserves naive image \textit{gradients}, \ie pixels with high Euclidean differences in the color channels.

\begin{figure}
\includegraphics[width=\linewidth]{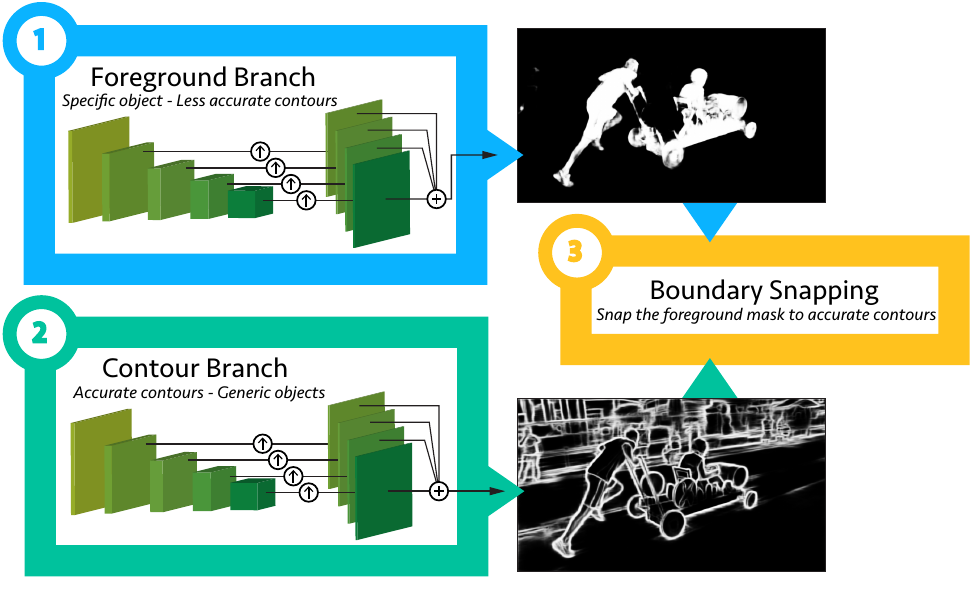}
\caption{\textbf{Two-stream FCN architecture}: The main foreground branch (1) is complemented by a contour branch (2) which improves the localization of the boundaries (3).}
\label{fig:two-stream}
\vspace{-2mm}
\end{figure}

To overcome this limitation, our second approach snaps the results to \textit{learned} contours instead of simple image gradients. To this end, we propose a complementary CNN in a second branch, that is trained to detect object contours. 
The proposed architecture is presented in Figure~\ref{fig:two-stream}: (1) shows the main foreground branch, where the foreground pixels are estimated; (2) shows the contour branch, which detects all contours in the scene (not only those of the foreground object).
This allows us to train offline, without the need to fine-tune on a specific example online.
We used the exact same architecture in the two branches, but training for different losses.
We noticed that jointly training a network with shared layers for both tasks rather degrades the obtained results thus we kept the computations for the two objectives uncorrelated.
This allows us to train the contour branch only offline and thus it does not affect the online timing.
Since there is need for high recall in the contours, we train on the PASCAL-Context~\cite{Mot+14} database, which provides contour annotations for the full scene of an image.
Finally, in the boundary snapping step (Figure~\ref{fig:two-stream} (3), we compute superpixels that align to the computed contours (2) by means of an Ultrametric Contour Map (UCM)~\cite{Arb+11,Pont-Tuset2016}, which we \textit{threshold} at a low value. We then take a foreground mask (1) and we select superpixels via majority voting (those that overlap with the foreground mask over 50\%) to form the final foreground segmentation.

In this second case, we trade accuracy for speed, since the snapping process takes longer (400 ms instead of 60 ms per frame), but we achieve more accurate results.
Both refinement processes result in a further boost in performance, and are fully modular, meaning that depending on the requirements one can choose not to use them, sacrificing accuracy for execution time, since both modules come with a small, yet avoidable computational overhead.

\section{Experimental Validation}
\label{sec:exp}
\paragraph*{Databases, state-of-the-art, and measures:}
The main part of our experiments is done on the recently-released DAVIS database~\cite{Perazzi2016},
which consists of 50 full-HD video sequences with all of their frames segmented with pixel-level accuracy. 
We use three measures: region similarity in terms of intersection over union ($\J$),
contour accuracy ($\F$), and temporal instability of the masks ($\T$).
All evaluation results are computed on the validation set of DAVIS.

We compare to a large set of state-of-the-art methods, including two very recent semi-supervised 
techniques, \nofl~\cite{\ofl}, \nbvs~\cite{\bvs}, as well as the methods originally compared on the DAVIS
benchmark: \nfcp~\cite{\fcp}, \njmp~\cite{\jmp}, \nhvs~\cite{\hvs}, \nsea~\cite{\sea}, and 
\ntsp~\cite{\tsp}.
We also add the unsupervised techniques: \nfst~\cite{\fst}, \nsal~\cite{\sal}, 
\nkey~\cite{\key}, \nmsg~\cite{\msg}, \ntrc~\cite{\trc}, \ncvos~\cite{\cvos}, and \nnlc~\cite{\nlc}.
We add two informative bounds: the quality that an oracle would reach by selecting the best segmented 
object proposal out of two state-of-the-art techniques (\ncob~\cite{\cob} and \nmcg~\cite{\mcg}), and by selecting the best superpixels from \ncob{} (COB$|$SP).

For completeness, we also experiment on Youtube-objects~\cite{Prest2012}, manually segmented by Jain and Grauman~\cite{Jain2014}. 
We compare to \nofl~\cite{\ofl}, \nbvs~\cite{\bvs}, \nltv~\cite{\ltv}, \nhbt~\cite{\hbt}, \nafs~\cite{\afs}, \nscf~\cite{\scf}, and \njfs~\cite{\jfs} and take the pre-computed evaluation results from previous work.
\vspace{-2mm}
\paragraph*{Ablation Study on DAVIS:}
To analyze and quantify the importance and need of each of the proposed blocks of our algorithm,
Table~\ref{tab:ablation} shows the evaluation of \ours{} compared to ablated versions without
each of its building blocks.
\begin{table}[b]
\setlength{\tabcolsep}{4pt} 
\centering
\footnotesize
\rowcolors{1}{white}{rowblue}
\resizebox{\linewidth}{!}{%
\sisetup{detect-all=true}
\begin{tabular}{llS[table-format=2.1]S[table-format=2.1]@{\hspace{1.5mm}}>{\fontsize{6}{6}}S[table-format=2.1]S[table-format=2.1]@{\hspace{1.5mm}}>{\fontsize{6}{6}}S[table-format=2.1]S[table-format=2.1]@{\hspace{1.5mm}}>{\fontsize{6}{6}}S[table-format=2.1]S[table-format=2.1]@{\hspace{1.5mm}}>{\fontsize{6}{6}}S[table-format=2.1]}
\toprule
\multicolumn{2}{c}{Measure} & \si{Ours} & \multicolumn{2}{c}{-\si{BS}} & \multicolumn{2}{c}{-\si{PN}-\si{BS}} & \multicolumn{2}{c}{-\si{OS}-\si{BS}} & \multicolumn{2}{c}{\scriptsize-\si{PN}-\si{OS}-\si{BS}} \\
\midrule                                 & Mean $\mathcal{M} \uparrow$     &\bfseries 79.8 &          77.4 &\itshape \color{blue}2.4 &          64.6 &\itshape \color{blue}15.2 &          52.5 &\itshape \color{blue}27.3 &          17.6 &\itshape \color{blue}62.2 \\
\cellcolor{rowblue}$\mathcal{J}$ & Recall $\mathcal{O} \uparrow$   &\bfseries 93.6 &          91.0 &\itshape \color{blue}2.6 &          70.5 &\itshape \color{blue}23.2 &          57.7 &\itshape \color{blue}35.9 &          2.3 &\itshape \color{blue}91.3 \\
                                 & Decay $\mathcal{D} \downarrow$  & 14.9 &          17.4 &\itshape \color{blue}2.5 &          27.8 &\itshape \color{blue}13.0 &\bfseries -1.9 &\itshape \color{red}16.7 &          1.8 &\itshape \color{red}13.1 \\
\hline
                                 & Mean $\mathcal{M} \uparrow$     &\bfseries 80.6 &          78.1 &\itshape \color{blue}2.5 &          66.7 &\itshape \color{blue}13.9 &          47.7 &\itshape \color{blue}32.9 &          20.3 &\itshape \color{blue}60.4 \\
\cellcolor{white}  $\mathcal{F}$ & Recall $\mathcal{O} \uparrow$   &\bfseries 92.6 &          92.0 &\itshape \color{blue}0.6 &          74.4 &\itshape \color{blue}18.3 &          47.9 &\itshape \color{blue}44.7 &          2.4 &\itshape \color{blue}90.2 \\
                                 & Decay $\mathcal{D} \downarrow$  & 15.0 &          19.4 &\itshape \color{blue}4.5 &          26.4 &\itshape \color{blue}11.4 &\bfseries 0.6 &\itshape \color{red}14.3 &          2.4 &\itshape \color{red}12.6 \\
\hline
\cellcolor{rowblue}$\mathcal{T}$ & Mean $\mathcal{M} \downarrow$   & 37.6 &\bfseries 33.5 &\itshape \color{red}4.0 &          60.9 &\itshape \color{blue}23.3 &          53.8 &\itshape \color{blue}16.2 &          46.0 &\itshape \color{blue}8.4 \\
\bottomrule
\end{tabular}
}\\[1.5mm]
\caption{\textbf{Ablation study on DAVIS}: Comparison of \ours{} against downgraded versions 
without some of its components.}
\label{tab:ablation}
\end{table}%
\begin{table*}[t!]
\setlength{\tabcolsep}{4pt} 
\centering
\footnotesize
\rowcolors{5}{white}{rowblue}
\resizebox{\textwidth}{!}{%
\sisetup{detect-weight=true}
\begin{tabular}{llS[table-format=2.1]S[table-format=2.1]S[table-format=2.1]S[table-format=2.1]S[table-format=2.1]S[table-format=2.1]S[table-format=2.1]S[table-format=2.1]S[table-format=2.1]S[table-format=2.1]S[table-format=2.1]S[table-format=2.1]S[table-format=2.1]S[table-format=2.1]S[table-format=2.1]S[table-format=2.1]S[table-format=2.1]S[table-format=2.1]}
\toprule
	 & & \multicolumn{8}{c}{Semi-Supervised} & \multicolumn{7}{c}{Unsupervised} & \multicolumn{3}{c}{Bounds} \\
\cmidrule(lr){3-10} \cmidrule(lr){11-17} \cmidrule(lr){18-20}
\multicolumn{2}{c}{Measure} & \si{Ours} & \si{\nofl} & \si{\nbvs} & \si{\nfcp} & \si{\njmp} & \si{\nhvs} & \si{\nsea} & \si{\ntsp} & \si{\nfst} & \si{\nnlc} & \si{\nmsg} & \si{\nkey} & \si{\ncvos} & \si{\ntrc} & \si{\nsal} & \si{COB|SP} & \si{\ncob} & \si{\nmcg} \\
\cmidrule(lr){1-2} \cmidrule(lr){3-10} \cmidrule(lr){11-17} \cmidrule(lr){18-20}
                                 & Mean $\mathcal{M} \uparrow$     &\bfseries 79.8 &          68.0 &          60.0 &          58.4 &          57.0 &          54.6 &          50.4 &          31.9 &\bfseries 55.8 &          55.1 &          53.3 &          49.8 &          48.2 &          47.3 &          39.3 &\bfseries 86.5 &          79.3 &          70.7 \\
\cellcolor{rowblue}$\mathcal{J}$ & Recall $\mathcal{O} \uparrow$   &\bfseries 93.6 &          75.6 &          66.9 &          71.5 &          62.6 &          61.4 &          53.1 &          30.0 &\bfseries 64.9 &          55.8 &          61.6 &          59.1 &          54.0 &          49.3 &          30.0 &\bfseries 96.5 &          94.4 &          91.7 \\
                                 & Decay $\mathcal{D} \downarrow$  &          14.9 &          26.4 &          28.9 &\bfseries -2.0 &          39.4 &          23.6 &          36.4 &          38.1 &\bfseries -0.0 &          12.6 &          2.4 &          14.1 &          10.5 &          8.3 &          6.9 &          2.8 &          3.2 &\bfseries 1.3 \\
\hline
                                 & Mean $\mathcal{M} \uparrow$     &\bfseries 80.6 &          63.4 &          58.8 &          49.2 &          53.1 &          52.9 &          48.0 &          29.7 &          51.1 &\bfseries 52.3 &          50.8 &          42.7 &          44.7 &          44.1 &          34.4 &\bfseries 87.1 &          75.7 &          62.9 \\
\cellcolor{white}$\mathcal{F}$ & Recall $\mathcal{O} \uparrow$   &\bfseries 92.6 &          70.4 &          67.9 &          49.5 &          54.2 &          61.0 &          46.3 &          23.0 &          51.6 &          51.9 &\bfseries 60.0 &          37.5 &          52.6 &          43.6 &          15.4 &\bfseries 92.4 &          88.5 &          76.7 \\
                                 & Decay $\mathcal{D} \downarrow$  &          15.0 &          27.2 &          21.3 &\bfseries -1.1 &          38.4 &          22.7 &          34.5 &          35.7 &\bfseries 2.9 &          11.4 &          5.1 &          10.6 &          11.7 &          12.9 &          4.3 &          2.3 &          3.9 &\bfseries 1.9 \\
\hline
\cellcolor{rowblue}$\mathcal{T}$ & Mean $\mathcal{M} \downarrow$   &          37.6 &          21.7 &          34.5 &          29.6 &          15.3 &          35.0 &\bfseries 14.9 &          41.2 &          34.3 &          41.4 &          29.1 &          25.2 &\bfseries 24.4 &          37.6 &          64.1 &\bfseries 27.4 &          44.1 &          69.8 \\
\bottomrule
\end{tabular}
}
\vspace{1mm}
\caption{\label{tab:evaltable}\textbf{DAVIS Validation}: \ours{} versus the state of the art, and practical bounds.}
\vspace{-3mm}
\end{table*}%
Each column shows: the original method without boundary snapping (-BS), without pre-training the
parent network on DAVIS (-PN), or without performing the one-shot learning on the specific sequence (-OS).
In smaller and italic font we show the loss (in blue) or gain (in red) on each metric with respect 
to our final approach.

We can see that both the pre-training of the parent network and the one-shot learning play an important role (we lose $15.2$ and $27.3$ points in $\J$ without them, respectively).
Removing both, \ie, using the Imagenet raw CNN, the results in terms of segmentation ($\J\!=\!17.6\%$) are completely \textit{random}.
The boundary snapping adds $2.4$ points of improvement, and is faster than conventional methods such as adding a CRF on top of the segmentation~\cite{Che+15}.

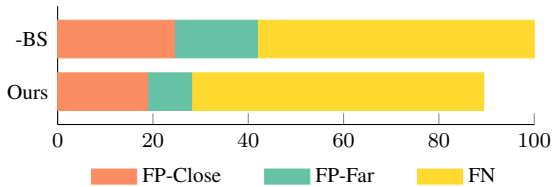
\begin{figure}[t]
\centering
\begin{tikzpicture}
\begin{axis}[
    xbar stacked,
    legend style={
    legend columns=3,
        at={(xticklabel cs:0.5)},
        anchor=north,
        draw=none
    },
    ytick=data,
    axis y line*=none,
    axis x line*=bottom,
    tick label style={font=\footnotesize},
    legend style={font=\footnotesize},
    label style={font=\footnotesize},
	legend style={/tikz/every even column/.append style={column sep=5mm}},
    xtick={0,20,40,60,80,100},
    width=0.95\linewidth,
    bar width=5mm,
    yticklabels={Ours,-BS},
    xmin=0,
    xmax=100,
    area legend,
    y=7mm,
    enlarge y limits={abs=0.625},
]
\addplot[Set2-8-2,fill=Set2-8-2] coordinates
{(18.87,0) (24.51,1) };
\addplot[Set2-8-1,fill=Set2-8-1] coordinates
{(9.24,0) (17.42,1) };
\addplot[Set2-8-6,fill=Set2-8-6] coordinates
{(61.3,0) (58.07,1) };
\legend{FP-Close, FP-Far, FN}
\end{axis}  
\end{tikzpicture}
\vspace{-1mm}
\caption{\textbf{Error analysis of our method}: Errors divided into False Positives (FP-Close and FP-Far) and False Negatives (FN). Values are total error pixels relative to the error in the -BS case.}
\label{fig:error_stats}
\end{figure}

Figure~\ref{fig:error_stats} further analyzes the type of errors that \ours{} produces (with and without boundary snapping), by dividing them into False Positives (FP) and False Negatives (FN).
FP are further divided into close and far, setting the division at 20 pixels from the object.
We can observe that the majority of the errors come from false negatives.
Boundary snapping mainly reduces the false positives, both the ones close to the boundaries (more accurate contours) and the spurious detections far from the object, because they do not align with the trained generic contours.

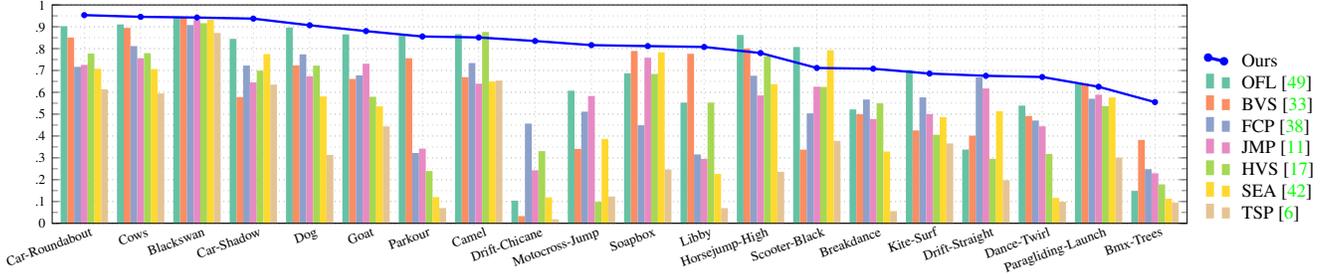
\begin{figure*}
\pgfplotstableread{data/per_seq_mean_J.txt}\perseqdata
\mbox{%
\begin{minipage}{0.91\textwidth}
  \resizebox{\textwidth}{!}{%
    \begin{tikzpicture}
        \begin{axis}[set layers,width=1.4\textwidth,height=0.35\textwidth,
                ybar=0pt,bar width=0.12,
                grid=both,
				grid style=dotted,
                minor ytick={0,0.05,...,1.1},
    			ytick={0,0.1,...,1.1},
			    yticklabels={0,.1,.2,.3,.4,.5,.6,.7,.8,.9,1},
                ymin=0, ymax=1,
                xtick = data, x tick label style={rotate=20,anchor=north east,xshift=7pt,yshift=5pt},
                xticklabels from table={\perseqdata}{Seq},
                major x tick style = transparent,
                enlarge x limits=0.03,
                font=\scriptsize,
                ]
            \addplot[draw opacity=0,fill=Set2-8-1,mark=none,legend image post style={yshift=-0.1em}] table[x expr=\coordindex,y=OFL]{\perseqdata};
            \label{fig:perseq:ofl}
            \addplot[draw opacity=0,fill=Set2-8-2,mark=none,legend image post style={yshift=-0.1em}] table[x expr=\coordindex,y=BVS]{\perseqdata};
            \label{fig:perseq:bvs}
            \addplot[draw opacity=0,fill=Set2-8-3,mark=none,legend image post style={yshift=-0.1em}] table[x expr=\coordindex,y=FCP]{\perseqdata};
            \label{fig:perseq:fcp}
            \addplot[draw opacity=0,fill=Set2-8-4,mark=none,legend image post style={yshift=-0.1em}] table[x expr=\coordindex,y=JMP]{\perseqdata};
            \label{fig:perseq:jmp}
            \addplot[draw opacity=0,fill=Set2-8-5,mark=none,legend image post style={yshift=-0.1em}] table[x expr=\coordindex,y=HVS]{\perseqdata};
            \label{fig:perseq:hvs}
            \addplot[draw opacity=0,fill=Set2-8-6,mark=none,legend image post style={yshift=-0.1em}] table[x expr=\coordindex,y=SEA]{\perseqdata};
			\label{fig:perseq:sea}
			\addplot[draw opacity=0,fill=Set2-8-7,mark=none,legend image post style={yshift=-0.1em}] table[x expr=\coordindex,y=TSP]{\perseqdata};
			\label{fig:perseq:tsp}
            \addplot[blue,sharp plot,update limits=false,mark=*,mark size=1,line width=1.2pt, legend image post style={yshift=-0.4em}] table[x expr=\coordindex,y=OSVOS]{\perseqdata};
            \label{fig:perseq:osdl}
            
            \addplot[black,sharp plot,update limits=false] coordinates{(-0.5,0) (20.5,0)};
        \end{axis}
    \end{tikzpicture}
    }
    \end{minipage}
    \hspace{0mm}
    \begin{minipage}{0.07\textwidth}
    \scriptsize
    \begin{tabular}{@{}l@{\hspace{1.5mm}}l}
    \ref{fig:perseq:osdl}& Ours\\
    \ref{fig:perseq:ofl} &\nofl~\cite{\ofl}\\
    \ref{fig:perseq:bvs} &\nbvs~\cite{\bvs}\\
    \ref{fig:perseq:fcp} &\nfcp~\cite{\fcp}\\
    \ref{fig:perseq:jmp} &\njmp~\cite{\jmp}\\
    \ref{fig:perseq:hvs} &\nhvs~\cite{\hvs}\\
    \ref{fig:perseq:sea} &\nsea~\cite{\sea}\\
    \ref{fig:perseq:tsp} &\ntsp~\cite{\tsp}
   \end{tabular}
    \end{minipage}
    }
    \vspace{-1mm}
    \caption{\label{fig:perseq}\textbf{DAVIS Validation}: Per-sequence results of region similarity ($\J$).}
\end{figure*}

\begin{figure*}
\centering
\resizebox{0.98\textwidth}{!}{%
	  \setlength{\fboxsep}{0pt}
      \rotatebox{90}{\hspace{4.5mm}Drift-Chicane}
      \fbox{\includegraphics[width=0.3\textwidth]{ims/examples/drift-chicane/00000_gt.png}}   
      \fbox{\includegraphics[width=0.3\textwidth]{ims/examples/drift-chicane/00015_ours.png}}
      \fbox{\includegraphics[width=0.3\textwidth]{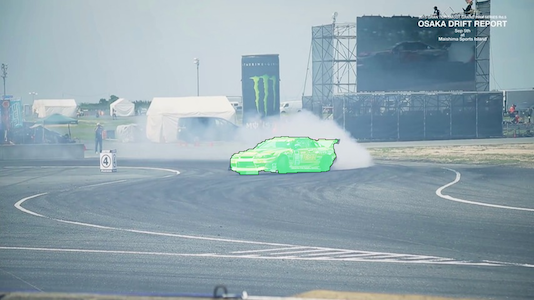}}
      \fbox{\includegraphics[width=0.3\textwidth]{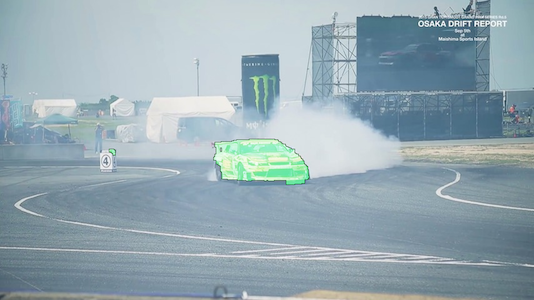}}
      \fbox{\includegraphics[width=0.3\textwidth]{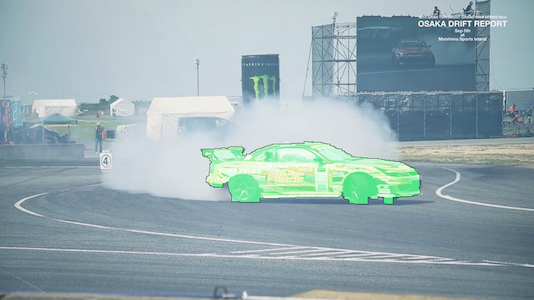}}
      }\\[1mm]
\resizebox{0.98\textwidth}{!}{%
	  \setlength{\fboxsep}{0pt}
      \rotatebox{90}{\hspace{6.5mm}Bmx-Trees}
      \fbox{\includegraphics[width=0.3\textwidth]{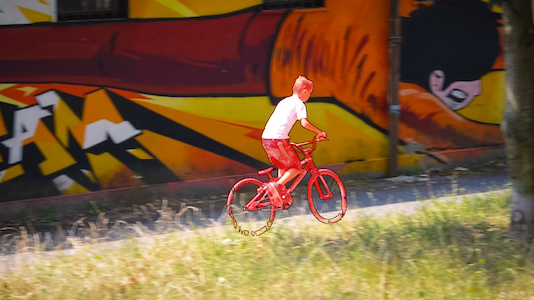}}
      \fbox{\includegraphics[width=0.3\textwidth]{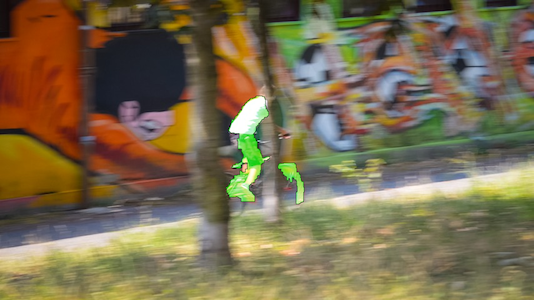}}
      \fbox{\includegraphics[width=0.3\textwidth]{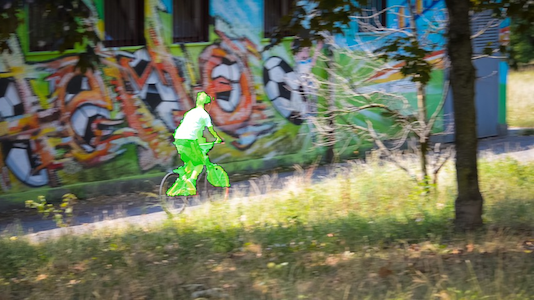}}
      \fbox{\includegraphics[width=0.3\textwidth]{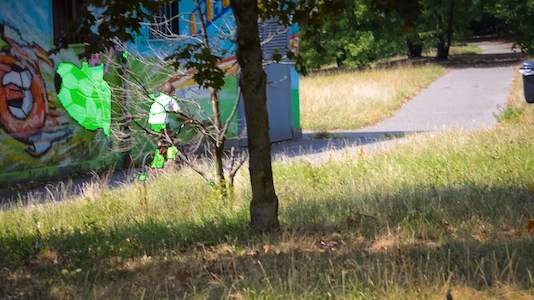}}
      \fbox{\includegraphics[width=0.3\textwidth]{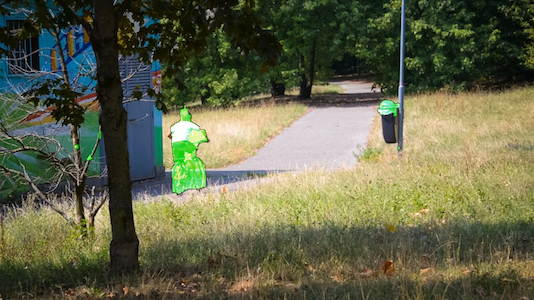}}
      }
\caption{\textbf{Qualitative results}: First row, an especially difficult sequence which \ours{} segments well. Second row, \ours{}' worst result.}
\label{fig:qualitative}
\vspace{-2mm}
\end{figure*}

\paragraph*{Comparison to the State of the Art on DAVIS:}
Table~\ref{tab:evaltable} compares \ours{} to the rest of the state of the art.
In terms of region similarity $\J$, \ours{} is $11.8$ points above the second best technique and
$19.8$ above the third best.
In terms of contour accuracy $\F$, \ours{} is $17.2$ and $21.8$ points above them.
Our results are better than those obtained by an oracle selecting
the best object proposal from the state-of-the-art object proposals \ncob.
Even if the oracle would select the best set of superpixels to form each mask
(COB$|$SP), \ours{} would be only $6.7$ points below.

Table~\ref{tab:evaltableattr} shows an evaluation with respect to different attributes annotated in the DAVIS dataset, by comparing the performance of the methods on the sequences with a given attribute (challenge) versus the performance on those without it.
\ours{} has the best performance on all attributes, and it has a significant resilience to these challenges (smallest decrease of performance when the attribute is present - numbers in italics).

\begin{table}[t]
\setlength{\tabcolsep}{4pt} 
\center
\footnotesize
\rowcolors{1}{white}{rowblue}
\resizebox{\linewidth}{!}{%
\sisetup{detect-all=true}
\begin{tabular}{lS[table-format=2.1]@{\hspace{1.5mm}}>{\fontsize{6}{6}}S[table-format=2.1]S[table-format=2.1]@{\hspace{1.5mm}}>{\fontsize{6}{6}}S[table-format=2.1]S[table-format=2.1]@{\hspace{1.5mm}}>{\fontsize{6}{6}}S[table-format=2.1]S[table-format=2.1]@{\hspace{1.5mm}}>{\fontsize{6}{6}}S[table-format=2.1]S[table-format=2.1]@{\hspace{1.5mm}}>{\fontsize{6}{6}}S[table-format=2.1]S[table-format=2.1]@{\hspace{1.5mm}}>{\fontsize{6}{6}}S[table-format=2.1]S[table-format=2.1]@{\hspace{1.5mm}}>{\fontsize{6}{6}}S[table-format=2.1]}
\toprule
Attr& \multicolumn{2}{c}{Ours} & \multicolumn{2}{c}{\nofl} & \multicolumn{2}{c}{\nbvs} & \multicolumn{2}{c}{\nfcp} & \multicolumn{2}{c}{\njmp} & \multicolumn{2}{c}{\nhvs} & \multicolumn{2}{c}{\nsea} \\
\midrule
AC	&\bfseries 80.6 &\itshape -1.2	&       56.6 &\itshape 17.6	&       48.6 &\itshape 17.6	&       52.8 &\itshape 8.6	&       52.4 &\itshape 7.0	&       41.4 &\itshape 20.4	&       43.2 &\itshape 11.1	\\
DB	&\bfseries 74.3 &\itshape 6.5	&       44.3 &\itshape 27.9	&       31.9 &\itshape 33.0	&       53.4 &\itshape 5.9	&       40.7 &\itshape 19.1	&       42.9 &\itshape 13.9	&       31.1 &\itshape 22.7	\\
FM	&\bfseries 76.5 &\itshape 5.1	&       49.6 &\itshape 28.2	&       44.8 &\itshape 23.3	&       50.7 &\itshape 11.9	&       45.2 &\itshape 18.0	&       34.5 &\itshape 31.0	&       30.9 &\itshape 30.1	\\
MB	&\bfseries 73.7 &\itshape 11.0	&       55.5 &\itshape 22.8	&       53.7 &\itshape 11.5	&       50.9 &\itshape 13.6	&       50.9 &\itshape 11.1	&       42.3 &\itshape 22.5	&       39.3 &\itshape 20.3	\\
OCC	&\bfseries 77.2 &\itshape 3.7	&       67.3 &\itshape 1.0	&       67.3 &\itshape -10.4	&       49.2 &\itshape 13.2	&       45.1 &\itshape 16.9	&       48.7 &\itshape 8.5	&       38.2 &\itshape 17.5	\\
\bottomrule
\end{tabular}}
\vspace{0.5mm}
\caption{\label{tab:evaltableattr} \textbf{Attribute-based performance}: Quality of the techniques on sequences with a certain attribute and the gain with respect to this quality in the sequences without it (in italics and smaller font). See DAVIS~\cite{Perazzi2016} for the meaning of the acronyms.
}
\vspace{-1mm}
\end{table}

Figure~\ref{fig:perseq} shows the results per sequence compared to the state of the art.
\ours{} has the best performance in the majority of sequences and is very close to the best in the rest.
The results are especially impressive in sequences such as Drift-Chicane or Bmx-Trees, where the majority
of techniques fail.
Figure~\ref{fig:qualitative} shows the qualitative results on these two sequences.
In the first row, the problem is especially challenging because of the smoke and the small initial size of the car.
In the second row, \ours{}' \textit{worse} sequence, despite vastly outperforming the rest of techniques. In this case, \ours{}
loses track of the biker when he is occluded, but recovers when he is visible again.
The rest of techniques lose the object because of the heavy occlusions.

\vspace{-3mm}
\paragraph*{Number of training images (parent network):}
To evaluate how much annotated data are needed to retrain a parent network, Table~\ref{tab:numtrainimages} shows the performance of OSVOS (-BS) when using a subset of the DAVIS train set. We randomly selected a fixed percentage of the annotated frames in each video.
\begin{table}[h]
\centering
\resizebox{0.75\linewidth}{!}{%
\rowcolors{2}{white}{rowblue}
\begin{tabular}{cccccccc}
\toprule
Training data  & 100 		& 200  		& 600 	   & 1000     & 2079  \\ \midrule
Quality ($\mathcal{J}$)  & 74.6	  	& 76.9    	&  77.2    & 77.3     & 77.4   \\
\bottomrule
\end{tabular}}
\vspace{2mm}
\caption{\textbf{Amount of training data}: Region similarity ($\J$) as a function of the number of training images. Full DAVIS is 2079.}
\label{tab:numtrainimages}
\end{table}
We conclude that by using only \texttildelow200 annotated frames, we are able to reach almost the same performance than when using the full DAVIS train split, thus not requiring full video annotations for the training procedure.
\vspace{-3mm}

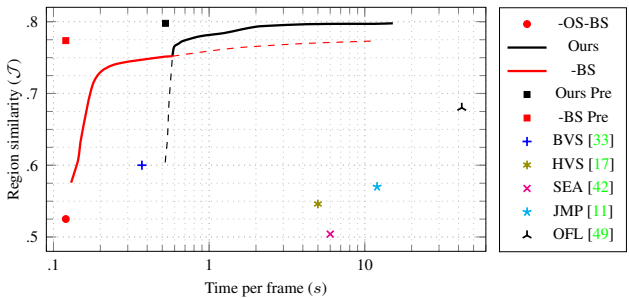
\begin{figure}[h]
\centering
\resizebox{\linewidth}{!}{\begin{tikzpicture}[/pgfplots/width=1\linewidth, /pgfplots/height=0.64\linewidth, /pgfplots/legend pos=south east]
    \begin{axis}[ymin=0.48,ymax=0.82,xmin=9e-2,xmax=60,enlargelimits=false,
        xlabel=Time per frame ($s$),
        ylabel=Region similarity ($\J$),
		font=\scriptsize,
        grid=both,
		grid style=dotted,
        xlabel shift={-2pt},
        ylabel shift={-5pt},
        xmode=log,
        legend columns=1,
        legend style={/tikz/every even column/.append style={column sep=3mm}},
        minor ytick={0,0.025,...,1.1},
        ytick={0,0.1,...,1.1},
		yticklabels={0,.1,.2,.3,.4,.5,.6,.7,.8,.9,1},
	    xticklabels={0,.1,1,10,100},
        legend pos= outer north east
        ]

		\addplot+[red,fill=red,mark=*, mark size=1.5,only marks, mark options={fill=red}] coordinates{(0.12,0.525)};
        \addlegendentry{-OS-BS}
        \label{fig:qual_vs_time:ours0}
        
        \addplot+[smooth,black,mark=none, line width=1] table[x=time,y=j_val] {data/quality_vs_time_ucm.txt};
        \addlegendentry{Ours}
        \label{fig:qual_vs_time:ours1lc}

        \addplot+[smooth,forget plot,dashed,black,mark=none, line width=0.5] table[x=time,y=j_val] {data/quality_vs_time_ucm_dashed.txt};

        
        \addplot+[smooth,red,mark=none, line width=1] table[x=time,y=j_val] {data/quality_vs_time_base.txt};
        \addlegendentry{-BS}
        \label{fig:qual_vs_time:ours1ns}
        
        \addplot+[smooth,forget plot,red,mark=none, line width=0.5, dashed] table[x=time,y=j_val] {data/quality_vs_time_base_dashed.txt};

        \addplot+[black,mark=square*, mark size=1.3,only marks, mark options={fill=black}] coordinates{(0.524469,0.797710)};
        \addlegendentry{Ours Pre}
        \label{fig:qual_vs_time:ours1lcpre}

        \addplot+[red,mark=square*, mark size=1.3,only marks, mark options={fill=red}] coordinates{(0.12,0.773520)};
        \addlegendentry{-BS Pre}
        \label{fig:qual_vs_time:ours1nspre}
        
        \addplot[blue,mark=+,only marks,line width=0.75] coordinates{(0.37,0.6)};
        \addlegendentry{\nbvs~\cite{\bvs}}
        \label{fig:qual_vs_time:bvs}
        
	    \addplot[olive,mark=asterisk, mark size=1.9,only marks, line width=0.75] coordinates{(5,0.546)};
        \addlegendentry{\nhvs~\cite{\hvs}}
        
        \addplot[magenta,mark=x, mark size=2.1,only marks, line width=0.75] coordinates{(6,0.504)};
        \addlegendentry{\nsea~\cite{\sea}}
        
        \addplot[cyan,mark=star, mark size=2,only marks, line width=0.75] coordinates{(12,0.57)};
        \addlegendentry{\njmp~\cite{\jmp}}

        \addplot[black,mark=Mercedes star, mark size=2.2,only marks, line width=0.75] coordinates{(42,0.68)};
        \addlegendentry{\nofl~\cite{\ofl}}
		\label{fig:qual_vs_time:ofl}
           
    \end{axis}
\end{tikzpicture}}
\vspace{-5mm}
   \caption{\textbf{Quality versus timing}: Region similarity with respect to the processing time per frame.}
   \label{fig:qual_vs_time}
   \vspace{-3mm}
\end{figure}

\paragraph*{Timing:}
The computational efficiency of video object segmentation is crucial for the algorithms to be usable in practice.
\ours{} can adapt to different timing requirements, providing progressively better results 
the more time we can afford, by letting the fine-tuning algorithm at test time do more or fewer iterations.
To show this behavior, Figure~\ref{fig:qual_vs_time} shows the quality of the result with respect to the time it takes to process each 480p frame.
As introduced before, \ours{}' time can be divided into the fine-tuning time plus the time to process each frame independently.
The first mode we evaluate is -OS-BS (\ref{fig:qual_vs_time:ours0}), in which we do not fine-tune to the particular sequence, and thus use the parent network directly. In this case, the quality is not very good (although comparable to some previous techniques), but we only need to do a forward pass of the CNN for each frame.

To take into account the fine-tuning time, we can consider two scenarios.
First, in Ours (\ref{fig:qual_vs_time:ours1lc}) or -BS (\ref{fig:qual_vs_time:ours1ns}) we average the fine-tuning time (done once per sequence) over the length of that sequence.
This way, the curves show the gain in quality with respect to the fine-tuning time, plus the forward pass on each frame. 
Using the same notation than in the ablation study, the two different curves refer to whether we do not perform boundary snapping (-BS) or we snap to the learned contours (Ours).
The better results come at the price of adding the snapping cost so depending on the needed speed, one of the two can be chosen.

Since \ours{} processes frames independently, one could also perform the fine-tuning offline, by  training on a picture of the object to be segmented beforehand (\eg take a picture of a racing horse before the race).
In this scenario, \ours{} can process each frame by one forward pass of the CNN (Ours Pre \ref{fig:qual_vs_time:ours1lcpre}, -BS Pre \ref{fig:qual_vs_time:ours1nspre}), and so be considerably fast.

Compared to other techniques, \ours{} is significantly faster and/or more accurate at all regimes, from fast modes: $74.7$ versus $60.0$ of \nbvs{} (\ref{fig:qual_vs_time:bvs}) at 400 ms, and $79.8$ versus $68.0$ of \nofl{} (\ref{fig:qual_vs_time:ofl}) at lower speeds.

\vspace{-3mm}
\paragraph*{Refinement of results:}
Another advantage of our technique is that we can naturally incorporate more supervision in the form of more annotated frames.
In a production environment, for instance, one needs a certain quality below which results are not usable.
In this scenario, \ours{} can provide the results with one annotated frame, then the operator can decide whether the quality is good enough, and if not, segment another frame. \ours{} can then incorporate that knowledge into further fine-tuning the result.

To model this scenario, we take the results with $N$ manual annotations, select the frame in which \ours{} performs worse, similarly to what an operator would do, \ie select a frame where the result is not satisfactory; and add the ground-truth annotation into the fine-tuning.
Table~\ref{tab:progressive} shows the evolution of the quality when more annotations are added (0 means we test the parent network directly, i.e.\ zero-shot).
We can see that the quality significantly increases from one to two annotations and 
saturates at around five.
As a measure of the upper bound of \ours{}, we fine-tuned on all annotated frames and 
tested on the same ones (last column), which indeed shows us that five annotated frames almost get the most out of this architecture.

\begin{table}[t]
\centering
\resizebox{0.95\linewidth}{!}{%
\rowcolors{2}{white}{rowblue}
\begin{tabular}{lccccccc}
\toprule
Annotations & 0 & 1 & 2 & 3 & 4 & 5 & All\\
\midrule
Quality ($\J$) & 58.5 & 79.8 & 84.6 & 85.9 & 86.9 & 87.5 & 88.7 \\
\bottomrule
\end{tabular}}
\vspace{2mm}
\caption{\textbf{Progressive refinement}: Quality achieved with respect to the number of annotated frames \ours{} trains from.}
\label{tab:progressive}
\end{table}

\begin{figure}[t]
\resizebox{\linewidth}{!}{%
\begin{minipage}{0.23\textwidth}
	  \setlength{\fboxsep}{0pt}
      \fbox{\includegraphics[width=\textwidth]{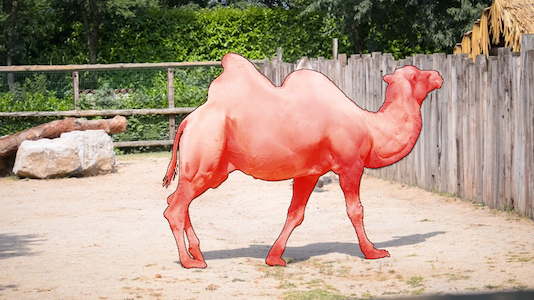}}\\[-1mm]
      \centering (a) Annotated frame 0
\end{minipage}\hspace{2mm}
\begin{minipage}{0.23\textwidth}
      \setlength{\fboxsep}{0pt}
      \fbox{\includegraphics[width=\textwidth]{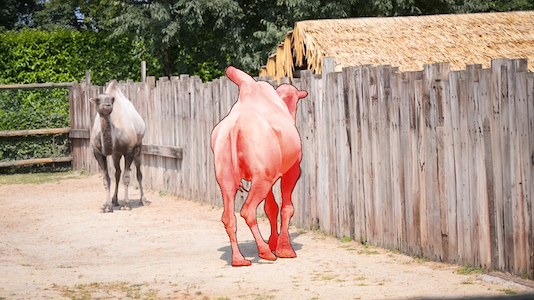}}\\[-1mm]
      \centering (c) Annotated frame 88
\end{minipage}\hspace{2mm}
\begin{minipage}{0.23\textwidth}
      \setlength{\fboxsep}{0pt}
      \fbox{\includegraphics[width=\textwidth]{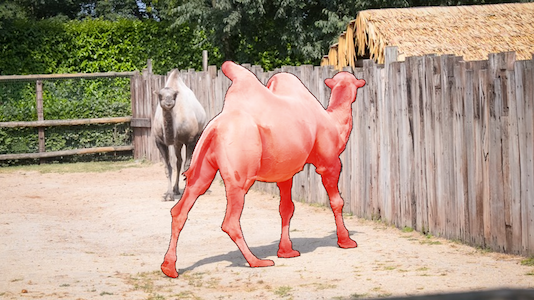}}\\[-1mm]
      \centering (e) Annotated frame 46
\end{minipage}    
      }\\[1mm]
\resizebox{\linewidth}{!}{%
\begin{minipage}{0.23\textwidth}
      \setlength{\fboxsep}{0pt}
      \fbox{\includegraphics[width=\textwidth]{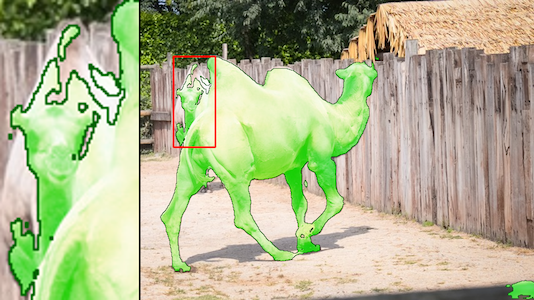}}\\[-1mm]
      \centering (b) Result frame 35
\end{minipage}\hspace{2mm}
\begin{minipage}{0.23\textwidth}
      \setlength{\fboxsep}{0pt}
      \fbox{\includegraphics[width=\textwidth]{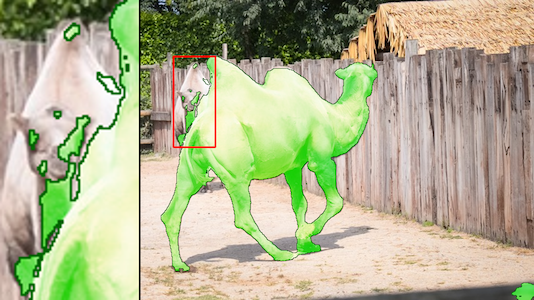}}\\[-1mm]
      \centering (d) Result frame 35
\end{minipage}\hspace{2mm}
\begin{minipage}{0.23\textwidth}
      \setlength{\fboxsep}{0pt}
      \fbox{\includegraphics[width=\textwidth]{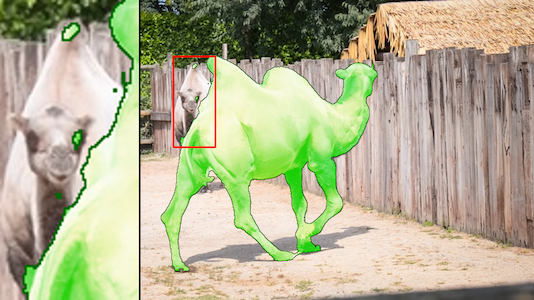}}\\[-1mm]
      \centering (f) Result frame 35
\end{minipage}    
      }
      \vspace{1mm}
\caption{\textbf{Qualitative incremental results}: The segmentation on frame 35 improves after frames 0, 88, and 46 are annotated.}
\label{fig:qual_increm}
\end{figure}

Figure~\ref{fig:qual_increm} shows a qualitative example of this process, where the user annotates frame 0, where only one camel is visible (a).
In frame 35, \ours{} also segments the second camel that appears (b), which has almost the exact same appearance.
This can be solved (f) by annotating two more frames, 88 (c) and 46 (e), which allows \ours{} to learn the difference between these two extremely similar objects, even without taking temporal consistency into account.

\vspace{-1mm}
\paragraph*{Evaluation as a tracker:}
Video object segmentation could also be evaluated as a Visual Object Tracking (VOT)~\cite{Kristan2015} algorithm, by computing the bounding box around each of the segmentations.
We compare to the winner of the VOT Challenge 2015~\cite{Kristan2015}: MDNET~\cite{Nam2016}.
Since we cannot compare in the original dataset of the VOT Challenge (the ground-truth objects are not segmented so we cannot fine-tune on it), we run MDNET on DAVIS. 
Table~\ref{tab:tracker} shows the percentage of bounding boxes coming from each technique that have an intersection over union with the ground-truth bounding box above different
thresholds. The higher the threshold, the more alignment with the ground truth is required.
We can see that \ours{} has significant better results as tracker than MDNET at all regimes, with more margin at higher thresholds.

\paragraph*{Results on Youtube-Objects:}
For completeness, we also do experiments on Youtube-objects~\cite{Prest2012,Jain2014}, where we take the pre-computed evaluation from other papers.
Table~\ref{tab:youtube} shows that we perform slightly better than the state of the art \nofl, which is significantly slower, and despite the fact that the sequences in this database have significant less occlusions and motion than in DAVIS, which favors techniques that enforce temporal consistency.

\begin{table}[t]
\centering
\resizebox{0.75\linewidth}{!}{%
\rowcolors{2}{white}{rowblue}
\begin{tabular}{lccccc}
\toprule
Overlap & 0.5 & 0.6 & 0.7 & 0.8 & 0.9\\
\midrule
Ours & \bf78.2 & \bf72.2 & \bf65.8 & \bf59.4 & \bf49.6 \\
MDNET~\cite{Nam2016}  & 66.4 & 57.8 & 43.4 & 29.5 & 14.7 \\
\bottomrule
\end{tabular}}
\vspace{2mm}
\caption{\textbf{Evaluation as a tracker}: Percentage of bounding boxes that match with the ground truth at different levels of overlap.}
\label{tab:tracker}
\end{table}

\begin{table}[t]
\centering
\rowcolors{2}{white}{rowblue}
\resizebox{1\linewidth}{!}{%
\begin{tabular}{lccccccccc}
\toprule
Category       & Ours       & \nofl      & \njfs      & \nbvs      & \nscf      & \nafs      & \nfst      & \nhbt      & \nltv      \\
\midrule
Aeroplane &    \ 88.2 & \bf\ 89.9 &    \ 89.0 &    \ 86.8 &    \ 86.3 &    \ 79.9 &    \ 70.9 &    \ 73.6 &    \ 13.7 \\
Bird      & \bf\ 85.7 &    \ 84.2 &    \ 81.6 &    \ 80.9 &    \ 81.0 &    \ 78.4 &    \ 70.6 &    \ 56.1 &    \ 12.2 \\
Boat      & \bf\ 77.5 &    \ 74.0 &    \ 74.2 &    \ 65.1 &    \ 68.6 &    \ 60.1 &    \ 42.5 &    \ 57.8 &    \ 10.8 \\
Car       &    \ 79.6 & \bf\ 80.9 &    \ 70.9 &    \ 68.7 &    \ 69.4 &    \ 64.4 &    \ 65.2 &    \ 33.9 &    \ 23.7 \\
Cat       & \bf\ 70.8 &    \ 68.3 &    \ 67.7 &    \ 55.9 &    \ 58.9 &    \ 50.4 &    \ 52.1 &    \ 30.5 &    \ 18.6 \\
Cow       &    \ 77.8 & \bf\ 79.8 &    \ 79.1 &    \ 69.9 &    \ 68.6 &    \ 65.7 &    \ 44.5 &    \ 41.8 &    \ 16.3 \\
Dog       & \bf\ 81.3 &    \ 76.6 &    \ 70.3 &    \ 68.5 &    \ 61.8 &    \ 54.2 &    \ 65.3 &    \ 36.8 &    \ 18.0 \\
Horse     & \bf\ 72.8 &    \ 72.6 &    \ 67.8 &    \ 58.9 &    \ 54.0 &    \ 50.8 &    \ 53.5 &    \ 44.3 &    \ 11.5 \\
Motorbike &    \ 73.5 & \bf\ 73.7 &    \ 61.5 &    \ 60.5 &    \ 60.9 &    \ 58.3 &    \ 44.2 &    \ 48.9 &    \ 10.6 \\
Train     &    \ 75.7 &    \ 76.3 & \bf\ 78.2 &    \ 65.2 &    \ 66.3 &    \ 62.4 &    \ 29.6 &    \ 39.2 &    \ 19.6 \\
\midrule
Mean      & \bf\ 78.3 &    \ 77.6 &    \ 74.0 &    \ 68.0 &    \ 67.6 &    \ 62.5 &    \ 53.8 &    \ 46.3 &    \ 15.5 \\
\bottomrule

\end{tabular}}
\vspace{2mm}
\caption{\textbf{Youtube-Objects evaluation}: Per-category mean intersection over union ($\J$).}
\label{tab:youtube}
\end{table}

\section{Conclusions}
\label{sec:concl}

Deep learning approaches often require a huge amount of training data in order to solve a specific problem such as segmenting an object in a video.  Quite in contrast, human observers can solve similar challenges with only a single training example.  In this paper, we demonstrate that one can reproduce this capacity of one-shot learning in a machine:  Based on a network architecture pre-trained on generic datasets, we propose \longours{} (\ours{}) as a method which fine-tunes it on merely one training sample and subsequently outperforms the state-of-the-art on DAVIS by 11.8 points.  Interestingly, our approach does not require explicit modeling of temporal consistency using optical flow algorithms or temporal smoothing and thus does not suffer from error propagation over time (drift). Instead, \ours{} processes each frame of the video independently and gives rise to highly accurate and temporally consistent segmentations. All resources of this paper can be found at \url{www.vision.ee.ethz.ch/~cvlsegmentation/osvos/}

\paragraph*{Acknowledgements:}
Research funded by the EU Framework Programme for Research and Innovation Horizon 2020 (Grant No. 645331, EurEyeCase), the Swiss Commission for Technology and Innovation (CTI, Grant No. 19015.1 PFES-ES, NeGeVA), and the ERC Consolidator Grant ``3D Reloaded''. The authors gratefully acknowledge support by armasuisse and thank NVidia Corporation for donating the GPUs used in this project.

{\small
\bibliographystyle{ieee}
\bibliography{cvpr2017}
}

\end{document}